\documentclass[11pt,a4paper]{article}
\usepackage[hyperref]{emnlp2018}
\usepackage{times}
\usepackage{latexsym}

\usepackage{url}

\usepackage{graphicx}
\usepackage{bm}
\usepackage{amsmath}
\usepackage{multirow}
\usepackage{rotating}

\aclfinalcopy % Uncomment this line for the final submission
\title{Neural Ranking Models for Temporal Dependency Structure Parsing}

\author{Yuchen Zhang \\
  Brandeis University \\
  {\tt yuchenz@brandeis.edu} \\\And
  Nianwen Xue \\
  Brandeis University \\
  {\tt xuen@brandeis.edu} \\}

\date{}

\begin{document}
\maketitle
\begin{abstract}
We design and build the first neural temporal dependency parser. It utilizes a neural ranking model with minimal feature engineering, and parses time expressions and events in a text into a temporal dependency tree structure. We evaluate our parser on two domains: news reports and narrative stories. In a parsing-only evaluation setup where gold time expressions and events are provided, our parser reaches 0.81 and 0.70 f-score on unlabeled and labeled parsing respectively, a result that is very competitive against alternative approaches. In an end-to-end evaluation setup where time expressions and events are automatically recognized, our parser beats two strong baselines on both data domains. Our experimental results and discussions shed light on the nature of temporal dependency structures in different domains and provide insights that we believe will be valuable to future research in this area.
\end{abstract}

\section{Introduction}

Temporal relation classification is important for  a range of NLP applications that include but are not limited to story timeline construction, question answering, summarization, etc. Most work on temporal information extraction models the task as a pair-wise classification problem \citep{bethard2007timelines,chambers07acl,chambers08emnlp,ning2018improving}: given an individual pair of time expressions and/or events, the system predicts whether they are temporally related and  which specific relation holds between them. An alternative approach is to model the temporal relations in a text  as a temporal dependency structure (TDS) for the entire text \citep{kolomiyets2012extracting}. Such a temporal dependency structure has the advantage that (1) it can be easily used to infer additional temporal relations between time expressions and/or events that are not directly connected via the transitivity properties of temporal relations, (2) it is computationally more efficient because a model does not need to consider all pairs of time expressions and events in a text, and (3) it is easier to use for downstream applications such as timeline construction. 

However, most existing automatic systems are pair-wise models trained with traditional statistical classifiers using a large number of manually crafted features \citep{tempeval2017}. 
%some more citations would be useful
The few exceptions include the work of \citet{kolomiyets2012extracting}, which describes a temporal dependency parser based on traditional feature-based classifiers, and \citet{dligach2017neural}, which describes a system using neural network based models to classify individual temporal relations. More recently, a semi-structured approach has also been proposed \citep{ning2018multi}.

In this work, taking advantage of a newly available data set annotated with temporal dependency structures -- the Temporal Dependency Tree (TDT) Corpus\footnote{\url{https://github.com/yuchenz/structured_temporal_relations_corpus}} \citep{zhang2018lrec}, we develop a neural temporal dependency structure parser using minimal hand-crafted linguistic features. One of the advantages of neural network based models  is that they are easily adaptable to new domains, and we demonstrate this advantage by evaluating our temporal dependency parser on data from two domains: news reports and narrative stories. Our results show that our model beats a strong logistic regression baseline. Direct comparison with existing models is impossible because the only similar dataset used in previous work that we are aware of is not available to us \citep{kolomiyets2012extracting}, but we show that our models are competitive against similar systems reported in the literature. 

The main contributions of this work are:
\vspace{-2mm}
\begin{itemize}
\setlength\itemsep{-5pt}
\item We design and build the first end-to-end neural temporal dependency parser. The parser is based on a novel neural ranking model that takes a raw text as input, extracts events and time expressions, and arranges them in a temporal dependency structure.
\item We evaluate the parser by performing experiments on data from two domains: news reports and narrative stories, and show that our parser is competitive against similar parsers. We also show the two domains have very different temporal structural patterns, an observation that we believe will be very valuable to future temporal parser development.
%\item We build an end-to-end system to evaluate how well a competitive temporal parser can perform on real-life data when both time expression and event recognition, and temporal relation extraction are automatically generated.
\end{itemize}

The rest of the paper is organized as follows.  Since temporal structure parsing is a relatively new task, we give a brief problem description in \S\ref{sec-problem-definition}. We describe our end-to-end pipeline system in \S\ref{sec-pipeline}. Details of the neural ranking model are discussed in \S\ref{sec-neural-model}. In \S\ref{sec-exp} we present and discuss our experimental results, and error analysis are presented in \S\ref{sec-error-analysis}. In \S\ref{sec-related-work} we discuss related work and situate our work in the broader context, and we conclude our paper in \S\ref{conclusion}.

\section{Problem Description}
\label{sec-problem-definition}
\begin{figure}
\centering
\includegraphics[scale=0.43]{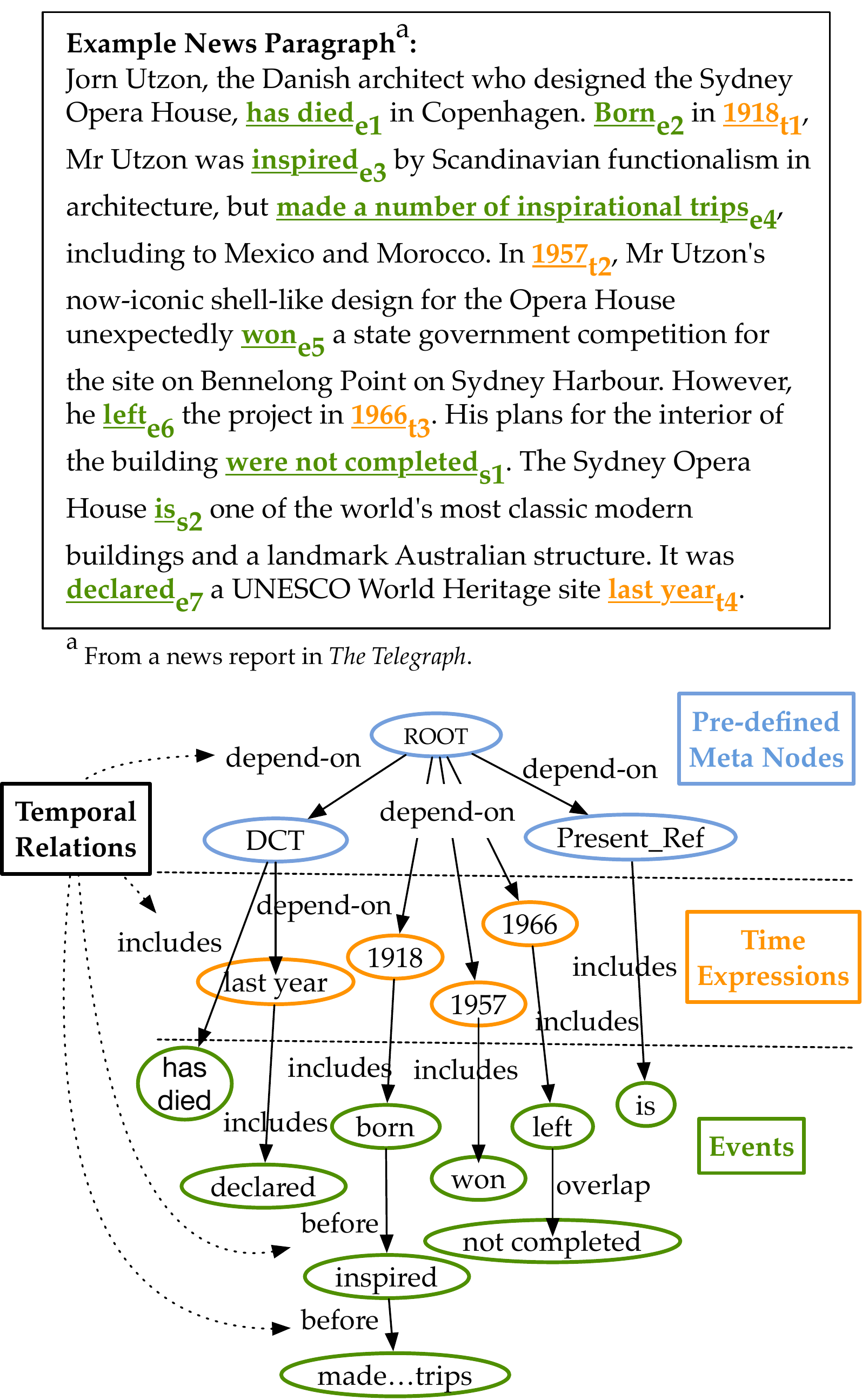}
\caption{Example text and its temporal dependency tree. DCT is Document Creation Time.}
\label{fig-tree-example}
\end{figure}

%A complete and formal definition of the temporal dependency structure for a text can be found in \citet{zhang2018lrec}. 
In this section we give a brief description of the temporal dependency parsing task (more details in \citet{zhang2018lrec}).
In a temporal structure parsing task, a text is parsed into a dependency tree structure that represents the inherent temporal relations among time expressions and events in the text. The nodes in this tree are mostly time expressions and events which are represented as contiguous spans of words in the text. They can also be  pre-defined meta nodes, which serve as reference times for  other time expressions and events, and they constitute the top-most part of the tree. For example, {\em Past\_Ref}, {\em Present\_Ref}, {\em Future\_Ref}, and {\em Document Creation Time (DCT)} are all pre-defined meta nodes. The edges in the tree represent temporal relations between each parent-child pair. The  temporal  relations can be one of {\em Includes}, {\em Before}, {\em Overlap}, and {\em After},  or  {\em Depend-on} which holds between two time expressions. Unlike syntactic dependency parsing where each word in a sentence is a node in the dependency structure, in a temporal dependency structure  only some of the words in a text are nodes in the structure. Therefore, this process naturally falls into two stages: first time expression and event recognition, and then temporal relation parsing. 
Figure~\ref{fig-tree-example} is an example temporal dependency tree for a news report paragraph.

Due to the fact that different types of time expressions and events behave differently in terms of what can be their antecedents, and recognition of these types can be helpful for determining temporal relations, finer classifications of time expressions and events are also defined. Time expressions are further classified into {\em Vague Time}, {\em Absolute Concrete Time}, and {\em Relative Concrete Time}, according to whether or not the time expression can be located on the timeline, and whether or not the interpretation of its temporal location depends on another time expression. Events are further classified into {\em Eventive Event}, {\em State}, {\em Habitual Event}, {\em Completed Event}, {\em Ongoing Event}, {\em Modalized Event}, {\em Generic Habitual}, and {\em Generic State}, according to the eventuality type of the event. Our  experiments will show that these fine-grained classifications are very helpful for the overall temporal structure parsing accuracy.

\section{A Pipeline System}
\label{sec-pipeline}
We build a two-stage pipeline system to tackle this temporal structure parsing problem. The first stage performs event and time expression identification. In this stage, given a text as input, spans of words that indicate events or time expressions are identified and categorized. We model this stage as a sequence labeling process. A standard Bi-LSTM sequence model coupled with BIO labels is applied here. Word representations are the concatenation of word and POS tag embeddings.

The second stage performs the actual temporal structure parsing by identifying the antecedent for each time expression and event, and identifying the temporal relation between them. In this stage, given events and time expressions identified in the first stage as input, the model outputs a temporal dependency tree in which each child node is an event or time expression that is temporally related to another event or time expression or pre-defined meta node as its parent node. This stage is modeled as a ranking process: for each node, a finite set of neighboring nodes are first selected as its candidate parents. These candidates are then ranked with a neural network model and the highest ranking candidate is selected as its parent. 
%An alternative approach is to use a transition-based or a graph-based model. 
We use a ranking model because it is simple, more intuitive and easier to train than a traditional transition-based or graph-based model, and the learned model rarely makes mistakes that violate the structural constraint of a tree. 

Since the model we use for Stage 1 is a very standard model with little modifications, we don't describe it in detail in this paper due to the limitation of space. Our neural ranking model for Stage 2 is described in detail in the next section.

\section{Neural Ranking Model}
\label{sec-neural-model}

\subsection{Model Description}
\label{sec-model-description}

We use a neural ranking model for the parsing stage.  For each time expression or event node $i$ in a text, a group of candidate parent nodes (time expressions, events, or pre-defined meta nodes) are selected. In practice, we select a window from the beginning of the text to two sentences after node $i$, and select all nodes in this window and all pre-defined meta nodes as the candidate parents if node $i$ is an event. Since the parent of a time expression can only be a pre-defined meta node or another time expression as described in \citet{zhang2018lrec}, we select all time expressions in the same window and all pre-defined meta nodes as the candidate parents if node $i$ is a time expression. Let $y'_i$ be a candidate parent of node $i$, a score is then computed for each pair of $(i, y'_i)$.%, which is the score for candidate $y'_i$ to be the parent of node $i$. 
Through ranking, the candidate with the highest score is then selected as the final parent for node $i$.

\begin{figure}
\centering
\includegraphics[scale=0.4]{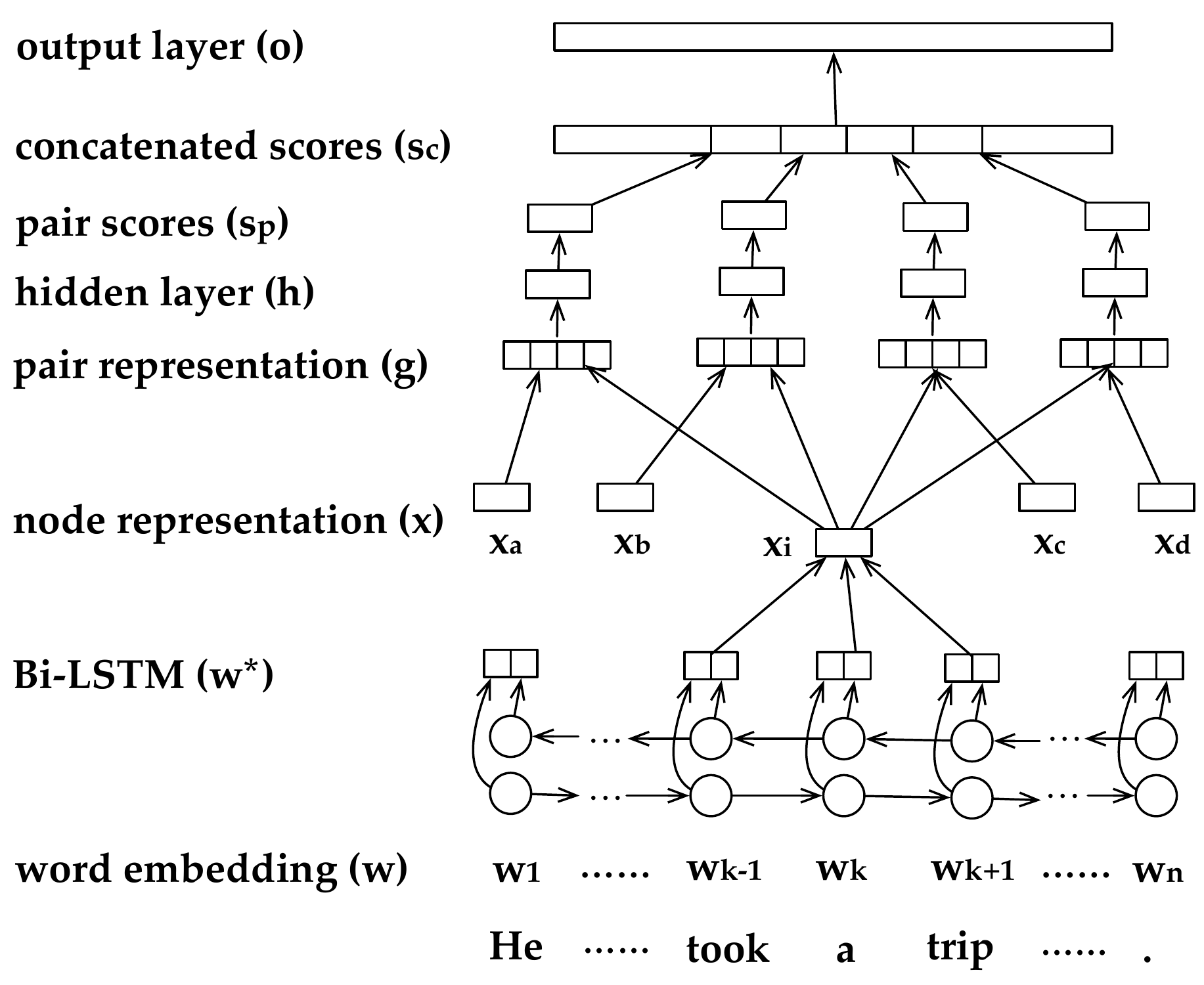}
\caption{Neural Ranking Model Architecture. $\bm{x_i}$ is the current child node, and $\bm{x_{a}, x_{b}, x_{c}, x_{d}}$ are the candidate parent nodes for $\bm{x_i}$. Arrows from Bi-LSTM layer to $\bm{x_{a}, x_{b}, x_{c}, x_{d}}$ are not shown.}
\label{fig-neural-model-architecture}
\end{figure}

Model architecture is shown in Figure~\ref{fig-neural-model-architecture}. Word embeddings are used as word representations (e.g. $\bm{w}_k$). A Bi-LSTM sequence layer is built on each word over the entire text, computing Bi-LSTM output vectors for each word (e.g. $\bm{w^*}_k$). The node representation for each time expression or event is the summation of the Bi-LSTM output vectors of all words in the text span (e.g. $\bm{x}_i$). The pair representation for node $i$ and one of its candidates $y'_i$ is the concatenation of the Bi-LSTM output vectors of these two nodes $\bm{g}_{i, y'_i} = [\bm{x}_i, \bm{x}_{y'_i}]$, which is then sent through a Multi-Layer Perceptron to compute a score for this pair ${s}_{i, y'_i}$. Finally all pair scores of the current node $i$ are concatenated into vector $\bm{c}_i$, and taking $softmax$ on it generates the final distribution  $\bm{o}_i$, which is the probability distribution of each candidate being the parent of node $i$.

Formally, the {\bf Forward Computation} is: 
\begin{align}
\bm{w^*}_k  & = BiLSTM(\bm{w}_k) \nonumber \\
\bm{x}_i & = sum(\bm{w^*}_{k-1}, \bm{w^*}_k, \bm{w^*}_{k+1}) \nonumber \\
\bm{g}_{i, y'_i}  & = [\bm{x}_i, \bm{x}_{y'_i}] \nonumber \\
\bm{h}_{i, y'_i}  & = tanh({\bf W_1} \cdot \bm{g}_{i, y'_i} + {\bf b_1}) \nonumber \\
{s}_{i, y'_i}  & = {\bf W_2} \cdot \bm{h}_{i, y'_i} + {\bf b_2}  \nonumber \\
\bm{c}_i & = [{s}_{i, 1},..., {s}_{i, i-1}, {s}_{i, i+1}, ..., {s}_{i, i+t}] \nonumber \\
\bm{o}_i &  = softmax(\bm{c}_i) \nonumber
\end{align}

\subsection{Learning}
Let D be the training data set of $K$ texts, $N_k$ the number of nodes in text $D_k$, and $y_i$ the gold parent for node $i$. Our neural model is trained to maximize $P(y_1, ..., y_{N_k} | D_k)$ over the whole training set. More specifically, the cost function is defined as follows:
\begin{align}
C & = -log \prod_{k}^{K} P(y_1, ..., y_{N_k}|D_k)  \nonumber\\
& = -log \prod_{k}^{K} \prod_{i}^{N_k} P(y_i|D_k)  \nonumber\\
& = \sum_{k}^{K} \sum_{i}^{N_k} - log P(y_i|D_k)  \nonumber
\end{align}

For each training example, cross-entropy loss is minimized:
\begin{align}
L & = -log P(y_i|D_k) \nonumber \\
& = -log \frac{exp[s_{i, y_i}]}{\sum_{y'_i} exp[s_{i, y'_i}]} \nonumber
\end{align}
where $s_{i, y'_i}$ is the score for child-candidate pair $(i, y'_i)$ as described in \S\ref{sec-model-description}.

% TO DO: can this model be trained in batches? I think so.

\subsection{Decoding}
During decoding, the parser constructs the temporal dependency tree incrementally by identifying the parent node for each event or time expression in textual order. To ensure the output parse is a valid dependency tree, two constraints are applied in the decoding process: (i) there can only be one parent for each node, and (ii) descendants of a node cannot be its parent to avoid cycles. Candidates violating these constraints are omitted from the ranking process.\footnote{An alternative decoding approach would be to perform a global search for a Maximum Spanning Tree. However, due to the nature of temporal structures, our greedy decoding process rarely hits the constraints.}

\subsection{Temporal Relation Labeling}
The neural model described above generates an unlabeled temporal dependency tree, with each parent being the most salient reference time for the child. However it doesn't model the specific temporal relation (e.g. ``before'', ``overlap'') between a parent and a child. We extend this basic architecture to both identify parent-child pairs and predict their temporal relations. In this new model, instead of ranking child-candidate pairs $(i, y'_i)$, we rank child-candidate-relation tuples $(i, y'_i, l_k)$, where $l_k$ is the $k$th relation in the pre-defined set of possible temporal relation labels $L$. We compute this ranking by re-defining the pair score $s_{i, y'_i}$. Here, pair score $s_{i, y'_i}$ is no longer a scalar score but a vector $\bm{s}_{i, y'_i}$ of size $|L|$, where $\bm{s}_{i, y'_i}[k]$ is the scalar score for $y'_i$ being the parent of $i$ with temporal relation $l_k$. Accordingly, the lengths of $\bm{c}_i$ and $\bm{o}_i$ are $number \ of \ candidates * |L|$. Finally, the tuple $(i, y'_i, l_k)$ associated with the highest score in $\bm{o}_i$ predicts that $y'_i$ is the parent for $i$ with temporal relation label $l_k$.

\subsection{Variations of the Basic Neural Model}
\subsubsection{Linguistically Enriched Models}
A variation of the basic neural model is a model that takes a few linguistic features as input explicitly. In this model, we extend the pair representation $\bm{g}_{i, y'_i}$ with local features: $\bm{g}_{i, y'_i} = [\bm{x}_i, \bm{x}_{y'_i}, \phi_{i, y'_i}]$. 

{\bf Time and event type feature:} Stage 1 of the pipeline not only extracts text spans that are time expressions or events, but also labels them with pre-defined categories of different types of time expressions and events. Readers are referred to \citet{zhang2018lrec} for the full category list. Through a careful examination of the data, we notice that  time expressions or events are selective as to what types of time expression or events can be their parent. In other words, the category of the child time expression or event has a strong indication on which candidate can be its  parent. For example, a time expression's parent can only be another time expression or a pre-defined meta node, and can never be an event; and an eventive event's parent is almost certainly another eventive event, and is highly unlikely to be a stative event. Therefore, we include the time expression and event type information predicted by stage 1 in this model as a feature. More formally, we represent a time/event type as a fixed-length embedding $\bm{t}$, and concatenate it to the pair representation $\bm{g}_{i, y'_i} = [\bm{x}_i, \bm{x}_{y'_i}, \bm{t}_i, \bm{t}_{y'_i}]$.
\vspace{2mm}

{\bf Distance features:} Distance information can be useful for predicting the parent of a child. Intuitively,  candidates that are closer to the child are more likely to be the actual parent. Through data examination, we also find that a high percentage of nodes have parents in close proximity. Therefore, we include two distance features in this model: the node distance between a candidate and the child $\bm{nd}_{i, y'_i}$, and whether they are in the same sentence $\bm{ss}_{i, y'_i}$. One-hot representations are used for both features to represent according conditions listed in Table~\ref{t-1hot-cond}.

\begin{table}[!h]
\small
\centering
\begin{tabular}{l}
\hline
conditions for feature $\bm{nd}_{i, y'_i}$: \\\hline
$i.node\_id - y'_i.node\_id = 1$ \\
$i.node\_id - y'_i.node\_id > 1$ and $i.sent\_id = y'_i.sent\_id$ \\
$i.node\_id - y'_i.node\_id > 1$ and $i.sent\_id \neq y'_i.sent\_id$ \\
$i.node\_id - y'_i.node\_id < 1$ \\\hline
conditions for feature $\bm{ss}_{i, y'_i}$: \\\hline
$i.sent\_id = y'_i.sent\_id$ \\
$i.sent\_id \neq y'_i.sent\_id$ \\\hline
\end{tabular}
\caption{\label{t-1hot-cond} Conditions for node distance and same sentence features. }
\end{table}

The final pair representation for our linguistically enriched model is as follows:
\begin{align}
\bm{g}_{i, y'_i} = [\bm{x}_i, \bm{x}_{y'_i}, \bm{t}_i, \bm{t}_{y'_i}, \bm{nd}_{i, y'_i}, \bm{ss}_{i, y'_i}] \nonumber
\end{align}

\subsubsection{Attention Model on Time and Event Representation}
In the basic neural model, a straight-forward sum-pooling is used as the multi-word time expression and event representation. However, multi-word event expressions usually have meaning-bearing head words. For example, in the event ``took a trip'', ``trip'' is more representative than ``took'' and ``a''. Therefore, we add an attention mechanism \citep{bahdanau2014neural} over the Bi-LSTM output vectors in each multi-word expression to learn a task-specific notion of headedness \citep{lee2017end}:
%Figure~\ref{fig-attention} illustrates the attention layer in the model architecture. 
\begin{align}
\alpha_t & = tanh(\bm{W} \cdot \bm{w}^*_{t}) \nonumber \\
w_{i, t} & = \frac{exp[\alpha_t]}{\sum^{END(i)}_{k=START(i)} exp[\alpha_k]} \nonumber \\
\hat{\bm{x}}_i & = \textstyle  \sum^{END(i)}_{t=START(i)} w_{i, t} \cdot \bm{w}^*_{t} \nonumber
\end{align}
where $\hat{\bm{x}}_i$ is a weighted sum of Bi-LSTM output vectors in span $i$. The weights $w_{i, t}$ are automatically learned. The final pair representation for our attention model is as follows:
\begin{align}
\bm{g}_{i, y'_i} = [\bm{x}_i, \bm{x}_{y'_i}, \bm{t}_i, \bm{t}_{y'_i}, \bm{nd}_{i, y'_i}, \bm{ss}_{i, y'_i}, \hat{\bm{x}}_i, \hat{\bm{x}}_{y'_i}] \nonumber
\end{align}

This model variation is also beneficial in an end-to-end system, where time expression and event spans are automatically extracted in Stage 1. When extracted spans are not guaranteed correct time expressions and events, an attention layer on a slightly larger context of an extracted span has a better chance of finding representative head words than a sum-pooling layer strictly on words within a event or time expression span. 

% TO DO: more features. (e.g. the pair representation ${\bf g}$ can be extended to include more features such as POS tags, distance embeddings, and time expression and event category labels. Word representation ${\bf x}$ can also be extended to include  character embeddings. %% Since the two stages in our pipeline both utilize a Bi-LSTM sequence model over the entire text, we can also experiment with allowing the two stages to share a Bi-LSTM model for multi-task Learning. )

\section{Experiments}
\label{sec-exp}
\subsection{Data}
All of our experiments are conducted on the datasets described in \citet{zhang2018lrec}. This is a temporal dependency structure corpus in Chinese. It covers two domains: news reports and narrative fairy tales. It consists of 115 news articles sampled from Chinese TempEval2 datasets \cite{tempeval2010} and Chinese Wikipedia News\footnote{\url{https://zh.wikinews.org}}, and 120 fairy tale stories sampled from Grimm Fairy Tales\footnote{\url{https://www.grimmstories.com/zh/grimm\_tonghua/index}}. 20\% of this corpus, distributed evenly on both domains, are double annotated with high inter-annotator agreements. We use this part of the data as our development and test datasets (10\% documents for development and 10\% for testing), and the remaining 80\% as our training dataset.

\subsection{Baseline Systems}
\label{sec-baseline}
We build two baseline systems to compare with our neural model. The first is a simple baseline which links every time expression or event to its immediate previous time expression or event. According to our data, if only position information is considered, the most likely parent for a child is its immediate previous time expression or event. This baseline uses the most common temporal relation edge label in the training datasets, i.e. ``overlap'' for news data, and ``before'' for grimm data. %This baseline doesn't rely on the first stage in the pipeline.

The second baseline is a more competitive baseline for stage 2 in the pipeline. It takes the output of the first stage as input, and uses a similar ranking architecture but with logistic regression classifiers instead of neural classifiers. The purpose of this baseline is to compare our neural models against a traditional statistical model under otherwise similar settings. We conduct robust feature engineering on this logistic regression model to make sure it is a strong benchmark to compete against. %Features we use include extensive time/event type features, distance features, lexical features, and various feature combinations. 
Table~\ref{t-log-reg-feat} lists the features and feature combinations used in this model. 

\begin{table}[!h]
\small
\centering
\begin{tabular}{l}
\hline
time type and event type features: \\\hline
$i.type$ and $y'_i.type$ \\
if $i.type = absolute \ time$ and $y'_i.type = root$ \\
if $i.type = time$ and $y'_i.type = root$ \\
are $i.type$ and $y'_i.type$ time, eventive, or stative\\
are $i.type$ and $y'_i.type$ root, time, or event \\
are $i.type$ and $y'_i.type$ root, time, eventive, or stative\\
if $i.type = y'_i.type = event$ and $\hat{y}.type = state,$ \\
\hspace{0.3in} for all $\hat{y}$ between $i$ and $y'_i$\\
\hline
distance features: \\\hline
if $i.sent\_id = y'_i.sent\_id$\\
 $i.node\_id - y'_i.node\_id$ \\
 if $i.node\_id - y'_i.node\_id = 1$\\\hline
combination features: \\\hline
if $i.type = state$ and $i.sent\_id \neq y'_i.sent\_id$ \\
if $i.type = state$ and $i.node\_id - y'_i.node\_id = 1$ \\
if $i.type = y'_i.type = event$ and \\
\hspace{0.3in} $i.node\_id - y'_i.node\_id = 1$ \\
if $i.type = state$ and $y'_i.type = event$ and \\
\hspace{0.3in} $i.node\_id - y'_i.node\_id = 1$ and \\
\hspace{0.3in} $i.node\_id\_in\_sent = 1$ and \\
\hspace{0.3in} $i.sent\_id \neq 1$ \\\hline
other features: \\\hline
if $i$ and $y'_i$ are in quotation marks \\
\hline
\end{tabular}
\caption{\label{t-log-reg-feat} Features in the logistic regression system.}
\end{table}

\subsection{Evaluation}
\label{sec-eval}
We perform two types of evaluations for our systems. First, we evaluate the stages of the pipeline and the entire pipeline, i.e. end-to-end systems where both time expression and event recognition, as well as temporal dependency structures are automatically predicted. Our models are compared against the two strong baselines described in \S\ref{sec-baseline}. These evaluations are described in \S\ref{sec-end2end-eval}.

The second evaluation focuses only on the temporal relation structure parsing part of our pipeline (i.e. Stage 2), using gold standard time expression and event spans and labels. Since most previous work on temporal relation identification use gold standard time expression and event spans, this evaluation gives us some sense of how our models perform against models reported in previous work even though a strict comparison is impossible because different data sets are used. These evaluations are described in \S\ref{sec-temporal-eval}.  

All neural networks in this work are implemented in Python with the DyNet library \citep{dynet}. The code is publicly available\footnote{\url{https://github.com/yuchenz/tdp_ranking}}.

%insert github url here
For Stage 1, all models are trained with Adam optimizer with early stopping and learning rate 0.001. The dimensions of word embeddings, POS tag embeddings, Bi-LSTM output vectors, and MLP hidden layers are tuned on the dev set to 256, 32, 256, and 256 respectively. POS tags in Stage 1 are acquired using the joint POS tagger from \citet{wang2014joint}. The tagger is trained on Chinese Treebank 7.0 \citep{xue2010chinese}. For Stage 2, the dimensions of word embeddings, time/event type embeddings, Bi-LSTM output vectors, and MLP hidden layers are tuned on the dev set to 32, 16, 32, and 32 respectively. The optimizer is Adam with early stopping and learning rate 0.001.

\begin{table}
\begin{center}
\begin{tabular}{c|ccc|ccc}
\hline
\bf evaluated  & \multicolumn{3}{c|}{\bf news} & \multicolumn{3}{c}{\bf grimm} \\
\bf label & \bf p & \bf r & \bf f & \bf p & \bf r & \bf f \\\hline
all span & .81 & .74 & .78 & .83 & .74 & .78 \\
 time & .83 & .81 & .82 & .97 & .62 & .76   \\
 event & .81 & .73 & .77 & .83 & .74 & .78  \\\hline
\end{tabular}
\end{center}
\caption{\label{t-stage1-result-coarse} Stage 1 cross-validation on span detection and binary time/event recognition.}
\end{table}

\begin{table}
\begin{center}
\begin{tabular}{ccccc}
\hline
{\bf time/event type}  & {\bf news} & {\bf grimm} \\\hline
vague time & .77 & .82 \\
concrete absolute & .67 & - \\
concrete relative  & .75 & - \\\hline
event  & .61 & .77 \\
state    & .65 & .61 \\
completed  & .62 & .26 \\
modalized &  .46 & .31 \\
%ongoing & .36 & .07 \\
%habitual event& 0 & 0\\
%generic habitual & 0 & 0\\
%generic state & 0 & 0\\
\hline
\end{tabular}
\end{center}
\caption{\label{t-stage1-result-fine} Stage 1 (time/event type recognition) cross-validation f1-scores on the full set.}
\end{table}

\subsubsection{End-to-End System Evaluation}
\label{sec-end2end-eval}

\paragraph{Stage 1: Time and Event Recognition}
\label{sec-stage1-eval}
For Stage 1 in the pipeline, we perform BIO tagging with the full set of time expression and event types (i.e. a 11-way classification on all extracted spans). Extracted spans will be nodes in the final dependency tree, and time/event types will support features in the next stage. We evaluate Stage 1 performance using 10-fold cross-validation of the entire data set. We use the ``exact match'' evaluation metrics for BIO sequence labeling tasks, and compute precision, recall, and f-score for each label type.

\begin{table*}
\centering
\begin{tabular}{c|c|cccc|cccc}
\hline
&\multirow{3}{*}{\bf model}& \multicolumn{4}{c|}{\bf news} & \multicolumn{4}{c}{\bf grimm} \\\cline{3-10}
&  & \multicolumn{2}{c}{\bf unlabeled f} & \multicolumn{2}{c|}{\bf labeled f} & \multicolumn{2}{c}{\bf unlabeled f} & \multicolumn{2}{c}{\bf labeled f} \\
& & \bf dev & \bf test &  \bf dev & \bf test & \bf dev & \bf test &  \bf dev & \bf test \\\hline
\multirow{5}{*}{\bf \begin{tabular}{@{}c@{}c@{}} temporal relation \\ parsing with \\ gold spans \end{tabular}} & Baseline-simple & .64 & .68 & .47 & .43 & .78 & {\bf.79} & .39 & .39 \\
&Baseline-logistic &  .81 & .79 & .63 & .54 & .74 & .74 & .60 & .63 \\
&Neural-basic  & .78 & .75 & .67 & .57 & .72 & .74 & .60 & .63 \\
&Neural-enriched & .80 & .78 & .67 & .59 & .76 & .77 & .63 & .65 \\
&Neural-attention & {\bf.83} & {\bf.81} & {\bf.76} & {\bf.70} & {\bf.79} & {\bf.79} & {\bf.66} & {\bf.68} \\\hline
\multirow{5}{*}{\bf \begin{tabular}{@{}c@{}c@{}} end-to-end \\ systems with \\ automatic spans \end{tabular}} & Baseline-simple & .39 & .40 & .26 & .25 & {\bf.44} & .47 & .27 & .25 \\
&Baseline-logistic & .36 & .34 & .24 & .22 & .43 & {\bf.49} & .33 & .37 \\
&Neural-basic  & .37 & .36 & .21 & .23 & .42 & .45 & .33 & .35 \\
&Neural-enriched & .51 & .52 & .32 & .35 & {\bf.44} & {\bf.49} & .33 & .37 \\
&Neural-attention & {\bf.54} & {\bf.54} & {\bf.36} & {\bf.39} & {\bf.44} & {\bf.49} & {\bf.35} & {\bf.39} \\\hline
\end{tabular}
\caption{\label{t-stage2-result} Stage 2 results (f-scores) with gold spans and timex/event labels (top), and automatic spans and timex/event labels generated by stage 1 (bottom). Best performances are in bold.}
\end{table*}

We first ignore fine-grained time/event types and only evaluate unlabeled span detection and time/event binary classification to show how well our system identify events and time expressions, and how well our system distinguishes time expressions from events. Table~\ref{t-stage1-result-coarse} shows the cross-validation results on these two evaluations. %Both span detection and binary time/event recognition have reached above $75\%$ f-score. 
Span detection and event recognition show similar performance on both news and narrative domains. Time expressions have a higher recognition rate than events in news data, which is consistent with the observation that time expressions usually have a more limited vocabulary and more strict lexical patterns. On the other hand, due to the scarcity of time expressions in the Grimm data, time expression recognition in this domain has a very high precision but low recall, which results in a much lower f-score than news.

Labeled full set evaluation results on time/event type classification are reported in Table~\ref{t-stage1-result-fine}. Time expressions have higher recognition rates than events on both domains, and dominant event types (``event'', ``state'', etc.) have  higher and more stable recognition rates than other types. Event types with very few training instances, such as ``modalized event'' ($<$7\%), achieve lower and more unstable recognition rates. Other types with less than $2$\% instances achieve close to 0 recognition f-scores, and are not reported in this table. 

\paragraph{Stage 2: Temporal Dependency Parsing}
\label{sec-stage2-eval}
% TO DO: add results and interpretation of the results on grimm data
For Stage 2 in the pipeline, we conduct experiments on the five systems described above: a simple baseline, a logistic regression baseline, a basic neural model, a linguistically enriched neural model, and an attention neural model. All models are trained on automatically predicted spans of time expressions and events, and time/event types generated by Stage 1 using 10-fold cross-validation, with gold standard edges (and edge labels) mapped onto the automatic spans. Evaluations in Stage 2 are against gold standard spans and edges, and evaluation metrics are precision, recall, and f-score on $\langle child, parent\rangle$ tuples for unlabeled trees, and $\langle child, relation, parent\rangle$ triples for labeled trees.  

Bottom rows in Table~\ref{t-stage2-result} report the end-to-end performance of our five systems on both domains. On both labeled and unlabeled parsing, our basic neural model with only lexical input performs comparable to the logistic regression model. And our enriched neural model with only three simple linguistic features  outperforms both the logistic regression model and the basic neural model on news, improving the performance by more than 10\%. However, our models only slightly improve the unlabeled parsing over the simple baseline on narrative Grimm data. This is probably due to (1) it is a very strong baseline to link every node to its immediate previous node, since in an narrative discourse linear temporal sequences are very common; and (2) most events breaking the temporal linearity in a narrative discourse are implicit stative descriptions which are harder to model with only lexical and distance features. Finally, attention mechanism improves temporal relation labeling on both domains. 

\subsubsection{Temporal Relation Evaluation}
\label{sec-temporal-eval}

To facilitate comparison with previous work where gold events are used as parser input, we report our results on temporal dependency parsing with gold time expression and event spans in Table~\ref{t-stage2-result} (top rows).
These results are in the same ballpark as what is reported in previous work on temporal relation extraction. %The best system reported in \citet{dligach2017neural} is a neural model trained and evaluated on Clinical TempEval data on pair-wise event-time ``contains'' relation, which reaches 0.70 unlabeled f-score. 
The best performance in \citet{kolomiyets2012extracting} are 0.84 and 0.65 f-scores for unlabeled and labeled parses, achieved by temporal structure parsers trained and evaluated on narrative children's stories. Our best performing model (Neural-attention) reports 0.81 and 0.70 f-scores on unlabeled and labeled parses respectively, showing similar performance. It is important to note, however, that these two works use different data sets, and are not directly comparable. Finally, parsing accuracy with gold time/event spans as input is substantially higher than that with predicted spans, showing the effects of error propagation.

\section{Error Analysis}
\label{sec-error-analysis}
%All following error analysis is performed on the output of our best model (Neural-attention) on the development sets. 
We perform error analysis on the output of our best model (Neural-attention) on the development data sets. We focus on analyzing our neural ranking model (i.e. Stage 2), with gold time expression and event spans and labels as input.

First, we look at errors by the types of antecedents. 
%We group all node types into three categories: time expressions, events, and  root, and draw a confusion matrix as in Table~\ref{t-cm-parent-1}. As shown in the table, while most reference time parent types are correctly identified, the most common error is wrongly recognized event parents when the correct parent should be the root (19\% error rate in news data and 43\% in Grimm data).
Most events in both news and grimm data depend on their immediate previous event or time expression as their reference time parent. 71\% of the events in the news data set and 78\% of the events in the Grimm data  have the immediate previous node as their antecedent. The confusion matrix in Table~\ref{t-cm-parent-2} illustrates how strongly this bias affects our models. Our model learns the bias and incorrectly links around half of the events (47\% in news and 46\% in grimm) to their immediate previous node when the {\em correct} temporal dependency is further back in the text.

\iffalse
\begin{table}[!h]
\begin{center}
\begin{tabular}{c|ccc|c}
\hline
{\bf news} & {\bf time} & {\bf event} & {\bf root} & {\bf total} \\\hline
{\bf time} & 60 & 14 &  9  & 83 \\
{\bf event} & 8 & 360 &  2 & 370  \\
{\bf root} & 2 & 28 & 117 & 147 \\\hline
{\bf total} & 70 & 402 & 128 & 600 \\ \hline\hline
{\bf grimm} & {\bf time} & {\bf event} & {\bf root}  & {\bf total}  \\\hline
{\bf time} & 13 & 0 &  1  & 14 \\
{\bf event} & 1 & 1027  &  1 & 1029 \\
{\bf root} & 0 & 9 & 12 & 21 \\\hline
{\bf total} & 14 & 1036 & 14 & 1064 \\ \hline
\end{tabular}
\end{center}
\caption{\label{t-cm-parent-1} Parent node type confusion matrix. Rows are gold parent types and columns are automatically parsed parent types.}
\end{table}
\fi

\begin{table}[!h]
\begin{center}
\begin{tabular}{c|ccc|ccc}
\hline
 & \multicolumn{3}{c|}{\bf news} & \multicolumn{3}{c}{\bf grimm} \\
& {\bf pre} & {\bf far} & {\bf total} & {\bf pre} & {\bf far} & {\bf total} \\\hline
{\bf pre} & 317 & 11 & 328 & 750  & 60 & 810 \\
{\bf far} & 65 & 72 & 137 & 104 &  122 & 226  \\\hline
{\bf total} & 382 & 83 & 465 & 854 & 182 & 1036 \\\hline
\end{tabular}
\end{center}
\caption{\label{t-cm-parent-2} Parent node confusion matrix. Rows are gold parents and columns are automatically parsed parents. ``pre'' means the parent is the immediate previous node of the child event, ``far'' means the parent is further back from the child event.}
\end{table}

\begin{table}[!h]
\begin{center}
\begin{tabular}{c|ccccc|c}
\hline
{\bf news} & {\bf be} & {\bf af} & {\bf ov} & {\bf in} & {\bf de} & {\bf total} \\\hline
{\bf before} & 1 & 0 &  21  & 2 & 0  & 24 \\
{\bf after}  & 1 & 0 &  1 & 0 & 0 & 2 \\
{\bf overlap} & 1 & 0 & 295 &  0 & 0 & 296  \\
{\bf include} & 0 & 0 & 4 & 52 & 0 & 56 \\
{\bf depend-on} & 0 & 0 & 0 & 0 & 117 & 117 \\\hline
{\bf total} & 3 & 0 & 321 & 54 & 117 & 495 \\\hline\hline
{\bf grimm} & {\bf be} & {\bf af} & {\bf ov} & {\bf in} & {\bf de} & {\bf total} \\\hline
{\bf before} & 367 & 0 &  55  & 0 & 0  & 422 \\
{\bf after}  & 1 & 0 &  2 & 0 & 0 & 3 \\
{\bf overlap} & 74 & 1 & 314 &  0 & 0 & 389  \\
{\bf include} & 3 & 0 & 0 & 10 & 0 & 13 \\
{\bf depend-on} & 0 & 0 & 0 & 0 & 12 & 12 \\\hline
{\bf total} & 445 & 1 & 371 & 10 & 12 & 839 \\\hline
\end{tabular}
\end{center}
\caption{\label{t-cm-temporal-relation} Temporal relation confusion matrix. Rows are gold relation labels and columns are automatic relation labels. ``be, af, ov, in, de'' stand for ``before, after, overlap, include, and depend-on''.}
\end{table}

Second, we look at errors in temporal relation labels. Considering only correctly recognized parent-child pairs, we draw a confusion matrix as in Table~\ref{t-cm-temporal-relation}. 
%Our data shows  few {\em after} relations in both domains, explaining the difficulty identifying this relation.
Our data has very few {\em after} relations in both domains, which explains why the model has difficulty identifying this relation.
There are also very few {\em include} and {\em depend-on} relations in the Grimm data, however they are identified with a relatively high accuracy. This is probably because, according to the temporal dependency structure design \citep{zhang2018lrec}, these relations hold only between restricted pairs of parent and child: {\em include} requires a time expression parent and an event child, and {\em depend-on} requires that the parent be the rootf. %And with 
The main confusion among temporal relations is between {\em before} and {\em overlap}. In news data, with a high occurrence of {\em overlap} relations (60\% {\em overlap} and 5\% {\em before}), most {\em before} parents are wrongly recognized as {\em overlap}. Grimm data has a more balanced distribution of these two temporal relations (46\% {\em overlap} and 50\% {\em before}), however, 13\% {\em before} and 17\% {\em overlap} are wrongly labeled as the other.

\section{Related Work}
\label{sec-related-work}

\subsection{Related Work on Temporal Relation Modeling}
There is a significant amount of research on temporal relation extraction \citep{bethard2007timelines,bethard2013cleartk,chambers08emnlp,chambers2014dense,ning2018improving}. Most of the previous work models temporal relation extraction as pair-wise classification between individual pairs of events and/or time expressions. Some of the models also add a global reasoning step to local pair-wise classification, typically using Integer Linear Programming, to exploit the transitivity property of temporal relations \citep{chambers08emnlp}. Such a pair-wise classification approach is often dictated by the way the data is annotated. In most of the widely used temporal data sets, temporal relations between individual pairs of events and/or time expressions are annotated independently of one another \citep{pustejovsky2003timebank,chambers2014dense,styler2014temporal,o2016richer,mostafazadeh2016caters}.

Our work is most closely related to  that of \citet{kolomiyets2012extracting}, which also treats temporal relation modeling as temporal dependency structure parsing. % Most prior work 
However, their dependency structure, as described in \citet{bethard2012annotating}, is only over events, excluding time expressions which are an important source of temporal information, and it also excludes {\em states} (stative events), which makes the temporal dependency structure incomplete. Moreover, their corpus only consists of data in the narrative stories domain.
%and (3) lacks linguistically founded guidelines for annotators to determine the parent of an event.  
We instead choose to develop our model based on the data set described in \citet{zhang2018lrec}, which introduces a more comprehensive and linguistically grounded annotation scheme for temporal dependency structures. This structure includes both events and time expressions, and uses the linguistic notion of {\em temporal anaphora} to guide the annotation of the temporal dependency structure. Since in this temporal dependency structure each parent-child pair is considered to be an instance of temporal anaphora, the parent is also called the {\it antecedent} and the child is also referred to as the {\it anaphor}. Their corpus consists of data from two domains: news reports and narrative stories.

More recently, \citet{ning2018multi} proposed a semi-structured approach to model temporal relations in a text. Based on the observation that not all pairs of events have well-defined temporal relations, they propose a multi-axis representation in which well-defined temporal relations only hold between events on the same axis. The temporal relations between events in a text form multiple disconnected subgraphs.
%Therefore, they first structure all events in a text onto different axes, and then model temporal relations among events on the same axis pair-wisely. 
Like other work before them, their annotation scheme only covers events, to the exclusion of  time expressions. 

\subsection{Related Work on Neural Dependency Parsing}

Most prior work on neural dependency parsing is aimed at syntactic dependency parsing, i.e. parsing a sentence  into a dependency tree that represents the syntactic relations among the words. Recent work on dependency parsing typically uses transition-based or graph-based architectures combined with contextual vector representations learned with recurrent neural networks (e.g. Bi-LSTMs) \citep{kiperwasser2016simple}. 
%The one prior work on temporal dependency parsing \citep{kolomiyets2012extracting} also used transition-based and graph-based architectures but are based on feature-based traditional classifiers. 

Temporal dependency parsing is, however, different from syntactic dependency parsing. In temporal dependency parsing, for each event or time expression, there is more than one other event or time expression that can serve as its reference time, while the most closely related one is selected as the gold standard reference time parent. This naturally falls into a ranking process where all possible reference times are ranked and the best is selected.
%due to the nature of temporal dependency structures and some major differences between syntactic and temporal dependency parsing tasks (discussed in more details in \S\ref{sec-why-ranking}), we claim 
%that a ranking model is a better fitting architecture for temporal dependency parsing than traditional transition-based or graph-based models. Therefore, our work is focused on building a neural-ranking-model-based temporal dependency parser.
%Our intuition is that a ranking model that identifies the most likely parent event/time expression  for each 
%child  event/time expression is more appropriate. 
In this sense our neural ranking model for temporal dependency parsing is closely related to the neural ranking model for coreference resolution described in \citet{lee2017end},
%Ranking models are used extensively in the context of information retrieval \citep{}, coreference resolution \citep{lee2017end}, recommender systems \citep{}, etc. Temporal dependency parsing is most comparable with coreference resolution tasks, 
both of which extract related spans of words (entity mentions for coreference resolution, and  events or time expressions for temporal dependency parsing). However, our temporal dependency parsing model differs from Lee et al's coreference model in that the ranking model for coreference only needs to output the best candidate for each individual pairing and cluster all pairs that are coreferent to each other. 
%and the decoding process usually focuses on each individual ranking decision and doesn't need to guarantee a structured output. For example, some coreference resolution systems output unstructured clusters of mentions by picking the best coreferent antecedent for each anaphor. 
In contrast, our ranking model for temporal dependency parsing needs to rank not only the candidate antecedents but also the temporal relations between the antecedent and the anaphor. In addition, the model also adds connectivity and acyclic constraints in the decoding process to guarantee a tree-structured output.

\section{Conclusion and Future Work}
\label{conclusion}
In this paper, we present the first end-to-end neural temporal dependency parser. We evaluate the parser with both gold standard and automatically recognized time expressions and events. In both experimental settings, the parser outperforms two strong baselines and shows competitive results against prior temporal systems. 

Our experimental results show that the model performance drops significantly when automatically predicted event and time expressions are used as input instead of gold standard ones, indicating an error propagation problem. Therefore, in future work we plan to develop joint models that simultaneously extract events and time expressions, and parse their temporal dependency structure. 
%That would require a much larger data set than currently available, so the first step would be to develop creative techniques to quickly additional data annotated with temporal dependency structures.

%\section*{Acknowledgments}

\bibliography{emnlp2018,nsf-2015}
\bibliographystyle{acl_natbib_nourl}

\end{document}